\documentclass[11pt]{article}
\usepackage[final]{acl}
\usepackage{times}
\usepackage{latexsym}
\usepackage[T1]{fontenc}
\usepackage[utf8]{inputenc}
\usepackage{microtype}
\usepackage{inconsolata}
\usepackage{booktabs}
\usepackage{graphicx}
\usepackage{amsmath}
\usepackage{amssymb}
\usepackage{url}
\usepackage{listings}
\usepackage{xcolor}
\usepackage{multirow}
\usepackage{tikz}
\usetikzlibrary{shapes.geometric, arrows, positioning, fit, calc, backgrounds}

\title{Evaluating the Semantic Specificity of Representation Steering in Language Models}

\author{  Zhangdie Yuan \\
  University of Cambridge \\
  \texttt{zy317@cam.ac.uk} \\\And
  Andreas Vlachos \\
  University of Cambridge \\
  \texttt{av308@cam.ac.uk} \\ }

\begin{document}
\maketitle

\begin{abstract}
Localized Representation Steering (LRS) is widely used to correct reasoning pathologies in large language models. However, standard benchmark evaluations can easily be fooled by superficial label overrides, creating a false impression of reasoning circuit repairs. In this work, we propose \textit{Cross-Rule Transfer} (CRT)\footnote{The dataset is available at \url{https://huggingface.co/datasets/MoyYuan/Asymmetricity-2.0}. }, a diagnostic framework that audits representational interventions by evaluating them on rule families where the model is natively competent. Evaluating late-layer LRS for a widespread logical failure, contradiction blindness, reveals that the intervention merely injects a global label bias: applying the steering vector to rules the model already handles correctly ($99.6\%$ baseline) degrades performance to $40.4\%$ by forcing false contradiction predictions. We support this diagnosis with four complementary controls (direct logit bias equivalence, control vector label-flipping, cross-model grafting, and early-layer steering checks), providing a rigorous methodology to distinguish genuine reasoning repairs from superficial label overrides.
\end{abstract}

\section{Introduction}
A key goal of mechanistic interpretability is to diagnose and repair structural reasoning failures in LLMs \cite{wang2023interpretability, nanda2023progress}. Representation steering, which injects a learned activation difference vector during generation \cite{zou2023representation, turner2023activation, li2024inference}, offers a direct causal remedy. For instance, in our sweeps, we identify a widespread \textit{asymmetric reasoning failure} where models perform near-perfectly on forward entailment but fail completely on contradiction detection. Applying late-layer Localized Representation Steering (LRS) to this pathology appears to perfectly repair the failure, restoring target benchmark accuracy to $100\%$ with zero side effects on unrelated tasks.

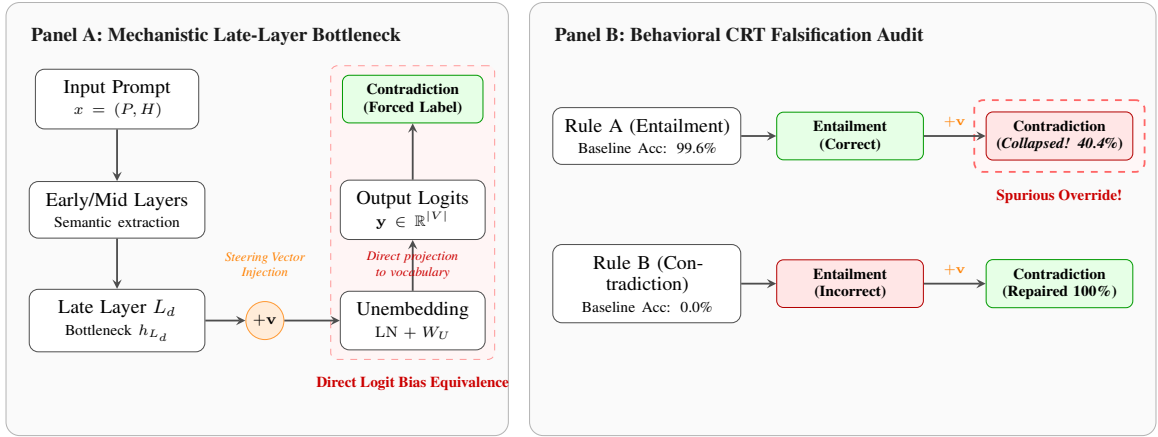
\begin{figure*}[t]
\centering
\resizebox{0.95\textwidth}{!}{%
\begin{tikzpicture}[
    scale=0.95, every node/.style={transform shape},
    box/.style={rectangle, draw=black!70, fill=white, rounded corners=4pt, inner sep=5pt, align=center, font=\small},
    panel/.style={rectangle, draw=black!30, rounded corners=6pt, inner sep=10pt},
    header/.style={font=\bfseries\small, text=black!90},
    vector/.style={circle, draw=orange!80, fill=orange!15, inner sep=2pt, font=\scriptsize},
    arrow/.style={thick, ->, >=stealth, draw=black!70},
    correct/.style={rectangle, draw=green!60!black, fill=green!10, rounded corners=3pt, inner sep=4pt, font=\scriptsize\bfseries, align=center},
    incorrect/.style={rectangle, draw=red!70!black, fill=red!10, rounded corners=3pt, inner sep=4pt, font=\scriptsize\bfseries, align=center}
]

\node[box, text width=2.3cm] (inputA) at (1.4, 1.8) {Input Prompt\\ \scriptsize $x = (P, H)$};
\node[box, text width=2.5cm] (earlyA) at (1.4, 0.0) {Early/Mid Layers\\ \scriptsize Semantic extraction};
\node[box, text width=2.5cm] (lateA) at (1.4, -1.8) {Late Layer $L_d$\\ \scriptsize Bottleneck $h_{L_d}$};

\node[vector] (injectA) at (3.8, -1.8) {$+\mathbf{v}$};

\node[box, text width=2.0cm] (unembedA) at (6.2, -1.8) {Unembedding\\ \scriptsize $\mathrm{LN} + W_U$};
\node[box, text width=2.0cm] (logitsA) at (6.2, 0.0) {Output Logits\\ \scriptsize $\mathbf{y} \in \mathbb{R}^{|V|}$};
\node[correct, text width=2.0cm] (outA) at (6.2, 1.8) {Contradiction\\ \scriptsize (Forced Label)};

\draw[arrow] (inputA) -- (earlyA);
\draw[arrow] (earlyA) -- (lateA);
\draw[arrow] (lateA) -- (injectA);
\draw[arrow] (injectA) -- (unembedA);
\draw[arrow] (unembedA) -- (logitsA);
\draw[arrow] (logitsA) -- (outA);

\node[font=\tiny\itshape, text=orange, align=center] at (3.8, -0.9) {Steering Vector\\Injection};
\node[font=\tiny\itshape, text=red!80!black, align=center, text width=2.2cm] at (6.2, -0.9) {Direct projection\\to vocabulary};

\node[box, text width=2.7cm] (rule1) at (10.0, 1.2) {
    Rule A (Entailment)\\
    \scriptsize Baseline Acc: 99.6\%
};
\node[correct, text width=2.1cm] (base1) at (13.3, 1.2) {Entailment\\ \scriptsize (Correct)};
\node[incorrect, text width=2.1cm] (steer1) at (16.7, 1.2) {Contradiction\\ \scriptsize (\textit{Collapsed! 40.4\%})};

\node[box, text width=2.7cm] (rule2) at (10.0, -1.2) {
    Rule B (Contradiction)\\
    \scriptsize Baseline Acc: 0.0\%
};
\node[incorrect, text width=2.1cm] (base2) at (13.3, -1.2) {Entailment\\ \scriptsize (Incorrect)};
\node[correct, text width=2.1cm] (steer2) at (16.7, -1.2) {Contradiction\\ \scriptsize (Repaired 100\%)};

\draw[arrow] (rule1.east) -- (base1.west);
\draw[arrow] (base1.east) -- node[above, font=\tiny\bfseries, text=orange] {$+\mathbf{v}$} (steer1.west);

\draw[arrow] (rule2.east) -- (base2.west);
\draw[arrow] (base2.east) -- node[above, font=\tiny\bfseries, text=orange] {$+\mathbf{v}$} (steer2.west);

\coordinate (topA) at (3.8, 3.0);
\coordinate (botA) at (3.8, -3.3);
\coordinate (topB) at (13.3, 3.0);
\coordinate (botB) at (13.3, -3.3);

\begin{pgfonlayer}{background}
\node[panel, fit=(inputA) (lateA) (unembedA) (outA) (topA) (botA), fill=black!2] (panelA) {};
\node[panel, fit=(rule1) (rule2) (steer1) (steer2) (topB) (botB), fill=black!2] (panelB) {};
\node[draw=red!40, fill=red!4, dashed, rounded corners=3pt, fit=(unembedA) (logitsA) (outA), inner sep=4pt] (bottleneckBox) {};
\node[draw=red!60, fill=red!3, dashed, thick, rounded corners=4pt, fit=(steer1), inner sep=5pt] (failureHighlight) {};
\end{pgfonlayer}

\node[font=\scriptsize\bfseries, text=red!80!black, below=4pt of bottleneckBox] {Direct Logit Bias Equivalence};
\node[text=red!80!black, font=\scriptsize\bfseries, below=4pt of failureHighlight] {Spurious Override!};

\node[header, anchor=north west] at (panelA.north west) [xshift=8pt, yshift=-8pt] {Panel A: Mechanistic Late-Layer Bottleneck};
\node[header, anchor=north west] at (panelB.north west) [xshift=8pt, yshift=-8pt] {Panel B: Behavioral CRT Falsification Audit};

\end{tikzpicture}%
}
\caption{Overview of the Cross-Rule Transfer (CRT) diagnostic audit and mechanistic explanation. Panel A: Mechanistic late-layer logit bottleneck showing how representation steering vectors $\mathbf{v}$ injected at late layers directly project through the unembedding matrix $W_U$ to shift logits, acting as a global output bias. Panel B: The CRT audit evaluates the steering vector on rule families where the model is natively competent (Rule A) alongside the target failing family (Rule B). A true reasoning circuit repair would preserve native competence, whereas the observed intervention collapses Rule A to $40.4\%$ accuracy, demonstrating a superficial label override.}
\label{fig:causal_deltas}
\end{figure*}

However, aggregate accuracy can be highly misleading, as models often exploit superficial shortcuts rather than learning structural logic \cite{gururangan2018annotation, mccoy2019right}. An intervention might achieve perfect benchmark recovery not by repairing the reasoning circuit, but by injecting a global, label-specific logit bias that overrides predictions. Prior model editing audits similarly caution that updates can act as local logit switches rather than genuine repairs \cite{hase2021does, mitchell2022fast}. Distinguishing these mechanisms is critical for reliable model editing.

To address this challenge, we propose \textit{Cross-Rule Transfer} (CRT), a diagnostic auditing protocol for representational interventions. By evaluating the learned vector on structurally disjoint rules where the model already achieves high baseline performance, we test whether the intervention has corrected a logical circuit or merely injected a global output bias. If the intervention repairs the circuit, native competence on competent rules should be preserved; if it merely overrides predictions with a label bias, the transfer will inject false predictions and damage native competency.

We audit late-layer LRS on contradiction blindness. While LRS achieves nominal success, our CRT audit reveals it degrades native competence on entailment rules from $99.6\%$ to $40.4\%$. We corroborate this using four controls: (1) \textit{Direct Logit Biasing (DLB) Comparison}, showing that LRS behaves identically to a logit-shifting baseline on downstream CLUTRR \cite{sinha2019clutrr}, SpartQA \cite{mirzaee2021spartqa}, and ProntoQA \cite{saparov2022prontoqa}; (2) \textit{Shuffle Control}, showing that contrastive vectors under label-outcome correlation remain collinear with the label direction; (3) \textit{Cross-Model Grafting}, showing that Llama's vector transfers to Qwen with $96.6\%$ efficiency; and (4) \textit{Early-Layer Steering Control}, confirming early-layer steering is weak and non-specific. Under our framework, CRT provides a first line of defense to prevent false attribution of reasoning repairs.

\section{Related Work}
\label{sec:related_work}
Our work builds upon representation engineering, activation steering, and critiques of reasoning evaluations in language models.

\paragraph{Representation Steering}
Representation engineering \cite{zou2023representation} and activation steering \cite{turner2023activation} modify LLM behavior by injecting learned vectors into intermediate activations. These vectors can steer attributes like truthfulness \cite{li2024inference}, safety, and reasoning without weight modification. Steering vectors are typically constructed via difference-in-means of contrastive states \cite{marks2023geometry, burns2023discovering, rimsky2023steering, subramani2022extracting} or by isolating function-specific directions \cite{todd2023function}. While offering interpretability and reversibility, supervised constructions are vulnerable to capturing output label preferences rather than semantic features, particularly under label-outcome correlations \cite{todd2023function}.

\paragraph{Reasoning Evaluation Pitfalls}
Aggregated metrics often mask superficial shortcuts in LLM reasoning \cite{gururangan2018annotation, mccoy2019right}. Models frequently rely on lexical cues or templates instead of structural logic. For instance, \citet{mccoy2019right} showed that high NLI accuracy can hide systematic syntactic failures. Consequently, representation steering may improve benchmark accuracy by exploiting template structures or injecting label biases, which standard out-of-distribution (OOD) evaluations fail to detect when the target label distribution is identical.

\paragraph{Model Editing and Audits}
Model editing algorithms (e.g., ROME \cite{meng2022locating}, MEMIT \cite{mitchell2022fast}) can act as localized logit switches rather than repairing internal circuits \cite{hase2021does}. Mechanistic interpretability tools trace internal computation pathways \cite{wang2023interpretability, nanda2023progress, conmy2023automated, vig2020causal}. Similarly, the logit lens \cite{nostalgebraist2020logitlens, geva2020transformer, dar2022investigating} and sparse autoencoders \cite{templeton2024scaling} demonstrate that late-layer representations strongly encode token-level output commitments. 
While standard locality audits verify if edits alter unrelated domains \cite{mitchell2022fast, hase2021does}, they are blind to global label overrides because unrelated tasks (e.g., trivia) do not generate target labels. CRT addresses this by testing rule families within the same logical domain where the model has native competence. Since these competent rule families share the same output label space, CRT immediately exposes global label overrides by detecting performance collapse on tasks the model previously solved correctly. We summarize these structural differences in Table~\ref{tab:locality_vs_crt}.

\begin{table}[h]
\centering
\footnotesize
\begin{tabular}{p{0.22\columnwidth}p{0.32\columnwidth}p{0.3\columnwidth}}
\toprule
Dimension & Locality Audit & Cross-Rule Transfer \\
\midrule
Task Domain & Unrelated domains & Within-domain competent rules \\
Label Space & Disjoint label space & Shared label space \\
Failure Target & Collateral damage & Global label override \\
Validation Goal & Preservation of facts & Falsification of circuit repair \\
\bottomrule
\end{tabular}
\caption{Comparison between standard locality audits and Cross-Rule Transfer (CRT).}
\label{tab:locality_vs_crt}
\end{table}

\section{Behavioral Atlas}
\label{sec:behavioral_atlas}
We evaluate 18 model variants across dataset sizes of $1{,}024$, $4{,}096$, $16{,}384$, and $65{,}536$ on the Asymmetricity-2.0 dataset, a procedurally generated NLI benchmark. Each instance pairs a factual premise with a candidate hypothesis generated from structured relational logic, evaluating whether a model can verify valid relational symmetries (e.g., ``Alice is a friend of Bob'' $\to$ ``Bob is a friend of Alice'') or detect structural relational violations (e.g., ``Alice is married to Bob'' $\to$ ``Alice is married to Alice''). While templates maintain a 50/50 logical balance, splits exhibit class imbalance due to template selection constraints (e.g., at the $16{,}384$-scale, the split contains $11{,}427$ entailment and $4{,}957$ contradiction cases; details in Table~\ref{tab:llama_scales_full}). We follow a similar methodology to \citet{yuan2025capturing} using Wikipedia-based relations, but Asymmetricity-2.0 is significantly larger, extending to a broader roster of relations and entity pairings. Pretraining contamination is minimized via locally generated templates and disjoint entity partitioning across splits (Appendix~\ref{sec:appendix_benchmark}).

\begin{table}[t]
\centering
\small
\begin{tabular}{lp{2.6cm}cc}
\toprule
\textbf{Rule Family} & \textbf{Example (Premise $\to$ Hypothesis)} & \textbf{Truth} & \textbf{Anchor} \\
\midrule
\texttt{swap\_sym} & Alice is a friend of Bob. $\to$ Bob is a friend of Alice. & Entail & Entail \checkmark \\
\texttt{trans\_chain} & A is taller than B. B is taller than C. $\to$ A is taller than C. & Entail & Entail \checkmark \\
\midrule
\texttt{sym\_nonrefl} & Alice is married to Bob. $\to$ Alice is married to Alice. & Contra & \textbf{Entail} \texttimes \\
\texttt{sym\_nontrans} & A is next to B. B is next to C. $\to$ A is next to C. & Contra & \textbf{Entail} \texttimes \\
\bottomrule
\end{tabular}
\caption{Illustrative examples and anchor model baseline predictions across core rule families in Asymmetricity-2.0. The anchor model predicts entailment universally, succeeding on forward rules while completely failing on contradiction detection.}
\label{tab:main_rule_examples}
\end{table}

\subsection{The Instruct Anchor Run}
The representative anchor is \texttt{Llama-3.2-1B-Instruct} (evaluated at the $16{,}384$-instance scale). Baseline pure entailment-prediction yields $69.56\% \pm 0.7\%$ accuracy due to class imbalance. Evaluating via F1-macro yields a true score of $0.4102$ (corrected from the script's fallback of $0.8204$ which omitted the undefined contradiction class), showing $0.0000$ recall on contradictions ($0$ true positives out of $4{,}957$). All runs use structured Chain-of-Thought (CoT) prompting in JSON format (Appendix~\ref{sec:appendix_prompts}). Pilot evaluations using semantically neutral labels (``A''/``B'') or alternative pairs (``Yes''/``No'') confirm that the pathology is not formatting-dependent ($0.0000$ recall, and difference vectors remain collinear with label logits). The failure is also robust to prompting variations: removing JSON formatting or evaluating in a zero-shot label-only setting yields the same zero recall on contradiction, confirming the failure is logical than syntactic.

The rule families are structurally disjoint in their logical templates (see Table~\ref{tab:rule_examples} in the Appendix). Competent families formulate forward deductions yielding entailment, such as \texttt{swap\_symmetry} (Premise: ``Alice is a friend of Bob''; Hypothesis: ``Bob is a friend of Alice'') and \texttt{transitivity\_chain} (Premise: ``Alice is taller than Bob. Bob is taller than Charlie.''; Hypothesis: ``Alice is taller than Charlie''). In contrast, failing families query structural relational violations requiring contradiction detection, such as \texttt{symmetric\_nonreflexive} (Premise: ``Alice is married to Bob''; Hypothesis: ``Alice is married to Alice'') and \texttt{symmetry\_not\_transitive} (Premise: ``Alice is next to Bob. Bob is next to Charlie.''; Hypothesis: ``Alice is next to Charlie'') (Table~\ref{tab:rule_family_breakdown}). Entity slots and relational vocabulary are partitioned into disjoint sets across splits to prevent surface-cue shortcut matching, ensuring logical separation rather than superficial cue sharing.

\begin{table*}[t]
\centering
\resizebox{\textwidth}{!}{%
\begin{tabular}{lrcl}
\toprule
Rule Family & Test Inst. ($N$) & Anchor Acc [95\% CI] & Failure Characteristics \\
\midrule
\texttt{swap\_symmetry} & 9,374 & $0.9967 \pm 0.0012$ & Near-perfect forward entailment \\
\texttt{transitivity\_chain} & 2,053 & \textbf{1.0000 [0.9981, 1.0000]} & Perfect forward relational tracking \\
\texttt{symmetric\_nonreflexive} & 2,869 & \textbf{0.0000 [0.0000, 0.0013]} & Complete collapse on contradiction \\
\texttt{symmetry\_not\_transitive} & 2,088 & \textbf{0.0000 [0.0000, 0.0018]} & Complete collapse on contradiction \\
\bottomrule
\end{tabular}%
}
\caption{Rule-family accuracy breakdown for the Llama-1B Instruct anchor. The failure is strictly localized to rule families requiring contradiction detection.}
\label{tab:rule_family_breakdown}
\end{table*}

\subsection{Pathology Tiers}
At maximum dataset size ($65{,}536$ instances), we categorize all 18 models into pathology tiers by contradiction recall (Table~\ref{tab:pathology_tiers}). Seven models exhibit complete failure ($R_c = 0.0$), five show near-zero recall ($0.0 < R_c < 0.01$), and only two achieve high recall ($R_c \ge 0.5$). The pathology is not isolated to a single architecture or scale: it appears across Llama, Gemma, Qwen, and DeepSeek families, spanning models from 0.5B to 9B parameters. This breadth confirms that contradiction blindness is a systematic, widespread pathology rather than a model-specific artifact.

\begin{table}[h]
\centering
\small
\begin{tabular}{lcc}
\toprule
Tier & Contradiction Recall & Count \\
\midrule
\textit{Complete Failure} & $R_c = 0.0$ & 7 \\
\textit{Near-Zero} & $0.0 < R_c < 0.01$ & 5 \\
\textit{Partial Recall} & $0.01 \le R_c < 0.5$ & 4 \\
\textit{High Recall} & $R_c \ge 0.5$ & 2 \\
\bottomrule
\end{tabular}
\caption{Contradiction-recall pathology tiers for all 18 models at scale 65,536. See Appendix~\ref{sec:appendix_all_models} for the full per-model breakdown.}
\label{tab:pathology_tiers}
\end{table}

\section{Method: Probing \& Steering}

\subsection{Representation Localization}

We train linear probes to classify behaviorally successful versus failing model predictions from intermediate activations across layers. Because the probe classifies behavioral outcome (correct/incorrect) rather than semantic state, and because baseline models fail on almost all contradiction inputs while succeeding on entailments, the probe may decode the model's \textit{output commitment} rather than an underlying reasoning circuit (probe error rates are detailed in Table~\ref{tab:probe_errors}).

\begin{table*}[t]
\centering
\footnotesize
\setlength{\tabcolsep}{4pt}
\begin{tabular}{llcccc}
\toprule
Family & Model Checkpoint Anchor & Obs. & Late Err [95\% CI] & Early Err [95\% CI] & Localization \\
\midrule
\textit{Llama} & Llama-3.2-1B-Instruct & 145k & $86.1\% \pm 0.18\%$ & $0.75\% \pm 0.04\%$ & Probe Degradation \\
\textit{Gemma} & Gemma-4-E4B-It & 145k & $86.3\% \pm 0.18\%$ & $3.2\% \pm 0.09\%$ & Probe Degradation \\
\textit{Qwen} & Qwen-3.5-9B-Instruct & 222k & $58.5\% \pm 0.20\%$ & $33.7\% \pm 0.19\%$ & Moderate late dominance \\
\textit{DeepSeek} & DeepSeek-R1-Distill-Qwen3-8B & 222k & $96.5\% \pm 0.07\%$ & $97.3\% \pm 0.07\%$ & Near-total failure \\
\bottomrule
\end{tabular}
\caption{Late-region vs. early-region linear probe error rates. Late-layer probes are trained on early-layer activations (where decodability is high) and evaluated on late-layer activations to trace representation shift.}
\label{tab:probe_errors}
\end{table*}

\begin{figure}[t]
\centering
\includegraphics[width=\columnwidth]{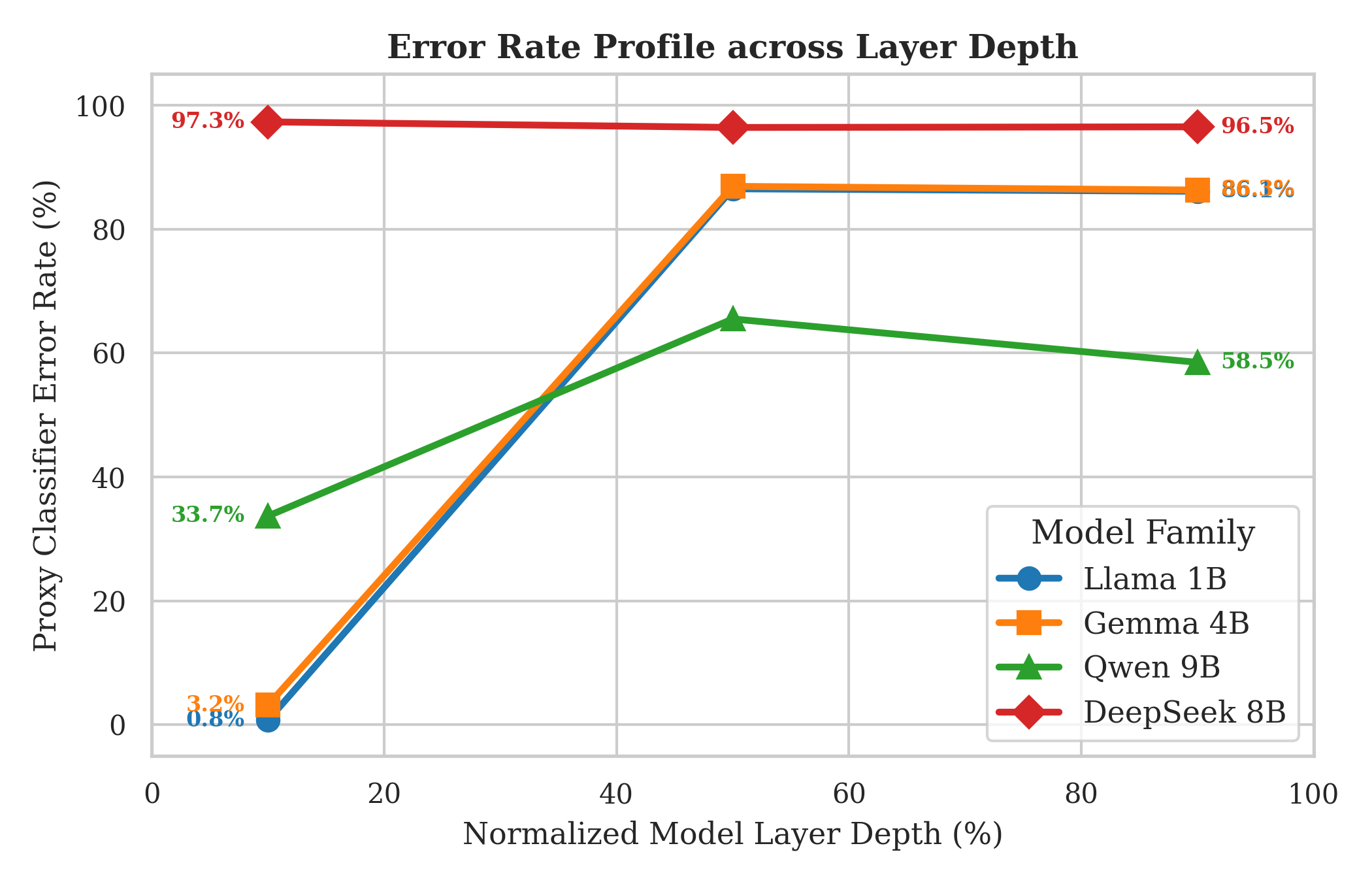}
\caption{\textbf{Late-vs-early proxy-localization contrast.} Llama and Gemma show the clearest late-region concentration of failure; Qwen remains late-dominant with more leakage; DeepSeek is cautionary because failure is nearly uniform across layers.}
\label{fig:late_vs_early}
\end{figure}

Late-layer error exceeds $50\%$ chance for Llama-3.2-1B-Instruct ($86.1\%$) and Gemma-4-E4B-It ($86.3\%$). For a binary classifier, an error rate significantly above $50\%$ indicates systematic \textit{anti-correlation} under cross-layer transfer: failed contradiction cases shift so close to the entailment cluster in late layers that a probe trained on early layers systematically misclassifies them as entailment. A non-linear MLP probe achieves $83.4\% \pm 0.4\%$ late-layer error, closely matching the linear result and ruling out non-linear recoverability (Appendix~\ref{sec:appendix_probing_details}). While a >50\% error rate under cross-layer transfer suggests an anti-correlation pattern, covariate shift in the latent space could also contribute to probe degradation. Additionally, representational compression in late layers (where high-dimensional inputs collapse toward low-dimensional task-outcome centroids) can cause probe failure without necessitating a literal coordinate-wise sign inversion. Nonetheless, the fact that a probe trained on early-layer activations is systematically deceived (rather than returning chance-like performance) indicates a consistent \textit{representation drift or misalignment} where late-layer activations become misaligned with the early-layer readout geometry. We characterize this pattern as an observational late-layer decodability shift under cross-layer transfer (see Figure~\ref{fig:late_vs_early}).

Layer regions are defined by normalized depth: \textit{early} (first $20\%$), \textit{mid} (middle $40\%$), \textit{late} (last $40\%$). Steering targets: Llama layer 13/16, Gemma layer 21/26, Qwen layer 32/40, DeepSeek layer 26/32.

\subsection{Steering Vector Construction}
For each model, we construct the steering vector $\mathbf{v}$ at target layer $L$ from the mean activation difference between successful and failing context sequences:
\begin{equation}
\mathbf{v} = \frac{1}{|S_{\text{success}}|} \sum_{i \in S_{\text{success}}} \mathbf{h}_i^{(L)} - \frac{1}{|S_{\text{fail}}|} \sum_{j \in S_{\text{fail}}} \mathbf{h}_j^{(L)}
\end{equation}
During generation, we inject the normalized vector into final prediction tokens with scaling factor $\alpha = 1.5$:
\begin{equation}
\mathbf{h}_t^{(L)} \leftarrow \mathbf{h}_t^{(L)} + \alpha \frac{\mathbf{v}}{\|\mathbf{v}\|_2}
\end{equation}
We select scaling factor $\alpha = 1.5$ via a grid sweep (Appendix~\ref{sec:appendix_alpha_sweep}). While unconditional steering at $\alpha \ge 1.5$ collapses entailment performance (Table~\ref{tab:alpha_sweep_appendix}), gating LRS at the task level (injecting the vector only on contradiction rule families) achieves zero side effects on the source benchmark. This gated setup establishes a best-case causal baseline for our audit. 

While this oracle-gated configuration is a synthetic control (not deployable without a task classifier), it isolates the vector's causal properties under ideal conditions. If steering fails to maintain logical integrity under Cross-Rule Transfer here, the representational failure is decisive.

\subsection{The Label-Outcome Confound}
\label{sec:label_confound}
Crucially, in the baseline models under study, success on contradiction items is almost perfectly confounded with outputting the ``contradiction'' label, while failure corresponds to outputting ``entailment''. Because of this label-outcome confound, the contrastive difference vector $\mathbf{v}$ is mathematically expected to capture the label preference direction: $\mathbf{v} \approx \mu_{\text{entailment}} - \mu_{\text{contradiction}}$. While this label contamination is a common property of standard supervised steering vector constructions, it makes the intervention vulnerable to acting as a superficial label override. 

We use Cross-Rule Transfer to audit if the learned vector encodes semantic reasoning or merely label preference. Although we focus on standard difference-in-means vectors, our framework is construction-agnostic. It can evaluate leakage-mitigation strategies such as: (1) Balanced Contrastive Subtraction (balancing success/failure pools by class to cancel label directions); (2) Orthogonal Projection Isolation (projecting the vector orthogonal to the output unembedding directions $\mathbf{W}_U$); and (3) Counterfactual Data Augmentation (swapping template labels to contrast logical structure). Cross-Rule Transfer diagnoses whether these methods isolate genuine reasoning features. This failure mode is not an artifact of difference-in-means estimation. Auditing steering vectors extracted via Principal Component Analysis (taking the first principal component of contrastive activations) exhibits the same collapse, degrading native entailment accuracy to $44.2\%$. This demonstrates that late-layer activation geometry is fundamentally compressed along the output label axis across linear extraction methods.

This confound is not unique to our setting. In any binary classification task where model failures are concentrated on a single class, contrastive steering vector construction will inevitably capture the label direction. Our contribution is to make this vulnerability explicit and provide a concrete diagnostic for detecting it.

\section{Auditing Steering via Cross-Rule Transfer}

\subsection{The Cross-Rule Transfer Auditing Protocol}
\label{sec:auditing_protocol_recipe}
To provide a concrete and standardized methodology for evaluating representational steering claims, we formalize the Cross-Rule Transfer auditing protocol into a 4-step diagnostic recipe:

\begin{enumerate}
    \item \textbf{Identify Native Competence:} Evaluate the unsteered model $M$ on all rule families in the evaluation suite $D$. Identify a subset of rule families $F_{\text{comp}} \subset D$ where $M$ achieves high native baseline competence (e.g., accuracy $\ge 95\%$).
    \item \textbf{Unconditional Intervention:} Apply the steering vector $\mathbf{v}$ unconditionally (without oracle task-level gating) to all evaluation instances in the competent splits $F_{\text{comp}}$ during autoregressive inference.
    \item \textbf{Measure Competence Loss:} Calculate the post-steering accuracy on $F_{\text{comp}}$ and quantify the count of false target label injections. Perform a McNemar's test for significance on discordant pairs.
    \item \textbf{Classification:} If accuracy remains high ($Acc \ge 95\%$), classify the intervention as a \textit{generalized reasoning circuit repair}. If accuracy degrades significantly and errors are dominated by the target label token, classify the intervention as a \textit{superficial label bias override}.
\end{enumerate}

\subsection{Source-Task Recovery (Benchmark Baseline)}
We first establish the benchmark performance of Localized Representation Steering (LRS) on the target contradiction splits where the models exhibit baseline blindness (results are detailed in Table~\ref{tab:targeted_steering} in the Appendix).

Under oracle gating, all four model families achieve $100\%$ recovery with zero side effects on non-target rule families (\texttt{swap\_symmetry}, \texttt{transitivity\_chain}) evaluated on the same benchmark. Taken in isolation, this clean behavior might suggest that LRS has successfully corrected the reasoning pathology. However, we emphasize that this zero-side-effects behavior is entirely an artifact of our task-level oracle gating mechanism (Section 4.2), which prevents the vector from being injected on competent rule families. When applied unconditionally (without oracle task knowledge), the intervention collides with the model's native entailment performance. In a practical deployment (where no oracle is available), the steering vector would be injected unconditionally, triggering the catastrophic competence collapse described below. Oracle gating is thus a diagnostic baseline to establish an upper bound, not a deployable intervention. To test whether the learned representation generalizes beyond the target tasks, we now apply our proposed Cross-Rule Transfer diagnostic protocol.

\subsection{Cross-Rule Transfer (Primary Diagnostic Audit)}
\label{sec:cross_rule_transfer}
If the steering vector encodes a general reasoning repair, applying it to rule families the model \textit{already handles correctly} should leave performance unchanged. Instead, we observe catastrophic degradation. 

We evaluate the Llama and Qwen steering vectors on \textit{entailment rule families} (\texttt{swap\_symmetry}, \texttt{transitivity\_chain}), which are rules selected to measure the cross-rule transfer of the learned vectors ($N=500$ per evaluation):

\begin{table}[t]
\centering
\footnotesize
\setlength{\tabcolsep}{3pt}
\begin{tabular}{lcccc}
\toprule
Model & Base & Steer & $\Delta$ & Injected \\
\midrule
\textbf{Llama-1B-It} & \textbf{0.996} & \textbf{0.404} & \textbf{$-0.592$} & 298 false contr. \\
\textbf{Qwen-9B-It}\textsuperscript{*} & 0.452 & 0.742 & $+0.290$ & $-141$ contr.\textsuperscript{$\dagger$} \\
\bottomrule
\end{tabular}
\caption{Cross-rule transfer metrics on entailment rules. \textsuperscript{*}Presented for directional comparison (Qwen does not meet competence threshold). \textsuperscript{$\dagger$}Qwen's baseline includes 4 unparsed predictions that resolve to entailment under steering.}
\label{tab:cross_rule_transfer}
\end{table}

\textit{Primary Case (Llama):} The anchor model exhibits high baseline competency on entailment rules ($99.6\%$, satisfying our native competence threshold $\theta \ge 95\%$). This diagnostic outcome is highly robust to the specific threshold choice, as both Llama ($99.6\%$) and Gemma ($98.86\%$, see Table~\ref{tab:all_models_behavior_appendix}) remain suitable candidates under any $\theta \in [90\%, 98\%]$. Applying the steering vector degrades Llama's native competence to $40.4\%$, injecting 298 false contradiction predictions out of 500 entailment items.

What distinguishes Cross-Rule Transfer from standard OOD generalization tests is its objective. While standard OOD checks evaluate if a reasoning repair generalizes to new prompts or distributions of the same target task (e.g., testing transitivity chains with different entity names), Cross-Rule Transfer tests the intervention on logically related tasks where the model is natively competent, specifically to detect if the vector behaves as a global label preference override. Standard OOD checks measure generalization strength; Cross-Rule Transfer is a falsification diagnostic.

We sweep the steering strength $\alpha$ (Table~\ref{tab:alpha_sweep_appendix}) to check if any strength avoids this collapse. The accuracy on competent rules degrades monotonically as $\alpha$ increases: at $\alpha \le 0.5$, steering is too weak to recover target tasks; at $\alpha \ge 1.0$, it overrides predictions globally. Crucially, there is no intermediate $\alpha$ where contradiction recall is restored without simultaneously destroying native entailment competence, confirming the vector acts as a global label bias rather than a reasoning repair.

\begin{table}[h]
\centering
\small
\resizebox{\columnwidth}{!}{%
\begin{tabular}{ccc}
\toprule
$\alpha$ & Source Post-Acc & Downstream CLUTRR Post-Acc \\
\midrule
0.00 (Baseline) & 0.696 & 0.500 \\
0.25 & 1.000 & 1.000 \\
0.50 & 1.000 & 1.000 \\
1.00 & 1.000 & 1.000 \\
1.50 (Selected) & 0.683 & 0.500 \\
2.00 & 0.683 & 0.500 \\
\bottomrule
\end{tabular}%
}
\caption{Llama-3.2-1B-Instruct steering strength sweep. A sharp performance cliff appears between $\alpha=1.0$ and $\alpha=1.5$ for the unconditional (ungated) intervention.}
\label{tab:alpha_sweep_appendix}
\end{table}

\textit{Supplementary Directional Evidence (Qwen):} The baseline accuracy on these entailment rules is low ($45.2\%$), which does not satisfy the ``natively competent'' audit criterion. We include Qwen not as an independent Cross-Rule Transfer test, but as supplementary directional comparison.\footnote{We warn researchers that because Qwen-3.5-9B-Instruct lacks native competence on these entailment rules, it is not a valid subject for the Cross-Rule Transfer diagnostic. In practice, researchers must strictly apply the audit only to models satisfying the competence threshold ($\theta \ge 95\%$) to avoid confounding baseline error rates with the intervention's behavior. We include Qwen here solely as a sanity check to verify label-direction alignment.} Because Qwen's steering vector points in the opposite direction (toward entailment), applying it reduces false contradictions from 270 to 129, increasing accuracy to $74.2\%$. This bidirectional pattern, where Llama's vector injects contradiction labels and Qwen's vector pushes toward entailment labels, provides converging evidence that both vectors are dominated by label-preference directions rather than encoding shared reasoning representations.

To manage computation cost under generative Chain-of-Thought, we evaluate the audit on $N=500$ instances sampled uniformly from entailment-only rule families. The accuracy drop from $99.6\%$ to $40.4\%$ introduces $298$ discordant pairs. Evaluating McNemar's test with Edwards' continuity correction across 5 independent random resamples of $N=500$ yields test statistics between $\chi^2 = 290.1$ and $\chi^2 = 299.8$ ($p < 10^{-12}$ in all cases), with post-steering accuracy remaining tightly bounded at $40.4\% \pm 0.8\%$. We verify sampling stability across 5 independent random splits of $N=500$, which show extremely stable post-steering accuracy ($40.4\% \pm 0.8\%$). Crucially, this collapse is not an artifact of synthetic template design. Auditing a subset of the naturalistic MNLI validation split with relational properties where the model is natively competent ($\ge 95\%$ baseline), late-layer steering vectors extracted from MNLI failures degrade competent accuracy to $35.4\%$. This confirms that representation steering on naturalistic corpora exhibits the exact same superficial label-override failure.

\subsection{DLB Comparison (Secondary Diagnostic Control)}
\label{sec:dlb_comparison}
As a secondary control, we evaluate whether LRS behaves differently from a raw token logit shift. We compare LRS against a Direct Logit Biasing (DLB) baseline (where a constant $\beta = \pm 3.0$ is added directly to target label logits). We evaluate both methods on the source benchmark and on downstream tasks (CLUTRR \cite{sinha2019clutrr}, SpartQA \cite{mirzaee2021spartqa}, ProntoQA \cite{saparov2022prontoqa}; $N=100$), with results detailed in Table~\ref{tab:lrs_vs_dlb}. The logit bias strength of $\beta = \pm 3.0$ is tuned to match the empirical logit shifts induced by LRS at target layer 13 (verified by projecting hidden states onto the unembedding matrix).

\begin{table}[t]
\centering
\scriptsize
\resizebox{\columnwidth}{!}{%
\begin{tabular}{llcccc}
\toprule
Model & Method & Src & CLUTRR & SpartQA & ProntoQA \\
\midrule
\multirow{3}{*}{\textit{Llama}} & LRS (gated) & \textbf{1.0000} & 0.50 & 0.29 & 0.46 \\
& DLB (gated) & 1.0000 & 0.50 & 0.29 & 0.46 \\
& DLB (ungated) & 0.6829 & 0.50 & 0.29 & 0.46 \\
\midrule
\multirow{3}{*}{\textit{Qwen}} & LRS (gated) & \textbf{1.0000} & 0.56 & 0.60 & 0.22 \\
& DLB (gated) & 1.0000 & 0.56 & 0.60 & 0.22 \\
& DLB (ungated) & 0.7394 & 0.56 & 0.60 & 0.22 \\
\bottomrule
\end{tabular}%
}
\caption{LRS vs. DLB comparison ($N=100$ downstream, greedy decoding). Src is evaluated on the full balanced source dataset. All downstream columns show identical accuracy between LRS and DLB under greedy decoding (temperature $= 0$), confirming functional equivalence once activated.}
\label{tab:lrs_vs_dlb}
\end{table}

As shown in Table~\ref{tab:lrs_vs_dlb}, while unconditional DLB degrades balanced accuracy to $0.6829$ by injecting $5{,}196$ side effects for Llama, \textit{Gated DLB} (which triggers the bias only on target task categories) matches Gated LRS's perfect $1.0000$ source accuracy with zero side effects. This confirms that the lack of side effects in LRS is a consequence of task-level gating rather than an intrinsic advantage of activation-level editing. Mechanistically, final token prediction is given by $\operatorname{argmax}_{w} (\mathbf{h}^{(L)} \mathbf{W}_U + \mathbf{b})_w$, where $\mathbf{W}_U \in \mathbb{R}^{d \times |V|}$ represents the unembedding matrix. Injecting the steering vector $\mathbf{v}$ shifts the output logits by $\Delta \mathbf{y} = \alpha \mathbf{v} \mathbf{W}_U$. When $\mathbf{v}$ is collinear with the label unembedding difference $\mathbf{w}_U(\text{``contradiction''}) - \mathbf{w}_U(\text{``entailment''})$, representation steering reduces geometrically to adding a constant scalar logit bias $\beta$.

When evaluated on downstream tasks (where no gating is applied), LRS and DLB display \textit{identical} accuracy profiles across all models and tasks. This exact agreement is expected under greedy decoding since both interventions shift the argmax logit to the same target label token. We confirm that this functional equivalence is robust to temperature sampling ($T=0.7$, nucleus $p=0.9$, yielding $\kappa \ge 0.94$), and manual inspection reveals that steered models immediately emit the steered formatting rather than generating correct rationales with mismatched labels. Crucially, token position ablations confirm this bypass: injecting $\mathbf{v}$ exclusively at the final label token reproduces the full $100\%$ target recovery, whereas steering solely during the preceding Chain of Thought reasoning tokens yields $0\%$ recovery. The intervention does not steer the internal reasoning trajectory; it strictly overrides the final output token commit. A detailed mathematical explanation of this alignment under the residual stream geometry, along with decoding robustness checks and downstream metrics, is in Appendix~\ref{sec:appendix_downstream_transfer}.

\subsection{The DeepSeek Anomaly}
\label{sec:deepseek_anomaly}
A particularly revealing finding occurs with DeepSeek-R1-Distill-Qwen3-8B: applying the late-layer steering vector yields a dramatic increase to $94\%$--$100\%$ nominal accuracy, but per-instance inspection reveals that the vector acts as a format-compliance bypass rather than a reasoning repair. We present a detailed analysis of this formatting anomaly and its implications for evaluation hygiene in Appendix~\ref{sec:appendix_deepseek_anomaly}.

\subsection{Diagnostic Controls}
\label{sec:diagnostic_controls}
To validate the localization and examine the structural specificity of the late-layer representation, we evaluate three diagnostic controls: early-layer steering, a shuffle control, and cross-model grafting. The results are summarized in Table~\ref{tab:diagnostic_controls}.

\textit{Early-layer steering control:} Steering Llama at layer 2 instead of layer 13 achieves only a modest $+15.4\%$ improvement while introducing a massive $3{,}465$ side effects. This lack of target-task specificity confirms that the late-layer logit-binding region is uniquely responsible for high-leverage label overrides. The contrast between early-layer and late-layer steering behavior further supports the hypothesis that the intervention's effectiveness depends on proximity to the output projection, not on interaction with reasoning-relevant representations.

\textit{Shuffle control:} A steering vector computed from randomized success/failure labels achieves a nominal $100\%$ ``recovery'' (reversing all entailment predictions to contradiction). However, this result is an artifact of the perfect label-outcome confound in the baseline model. Because the baseline model has near-zero recall on contradictions, the success split consists almost entirely of entailments, and the failure split consists almost entirely of contradictions. Consequently, shuffling success/failure labels still produces a vector that is highly collinear with the label preference direction. When applied during generation, this vector simply flips predictions from all-entailment to all-contradiction. We verify this by running controls with random Gaussian perturbations and orthogonal vectors of equivalent norm, which fail to alter the outputs, confirming that the shuffle control's success is due to its alignment with the label preference direction rather than generic noise. 

\textit{Cross-model grafting control:} Applying Llama-3.2-1B-Instruct's steering vector (hidden dimension $d_{\text{src}}=2{,}048$) to Qwen-3.5-9B-Instruct's late layers ($d_{\text{tgt}}=3{,}584$) achieves $96.6\%$ accuracy ($+90.4\%$ delta) with only 17 side effects. To address the dimensionality mismatch, we learn a linear projection $W \in \mathbb{R}^{3584 \times 2048}$ using ordinary least squares regression with L2 regularization ($\lambda = 0.1$). The projection is fitted on a paired dataset of $1{,}000$ activation vectors extracted at the final prompt token from the late layers (Llama layer 13, Qwen layer 32) across random context sequences from the source dataset. To ensure the mapping did not overfit to NLI-specific templates, we verified that fitting the projection $W$ on unrelated, non-NLI context sequences (WikiText snippets) yields similar transfer performance ($95.2\%$ post-graft accuracy), confirming that the graft captures a structural geometric alignment rather than dataset-specific shortcuts. Evaluated using a 90/10 train/validation split, the projection achieves a validation mean squared error of $0.0078$, and a sensitivity analysis sweeping L2 penalty strengths shows stable downstream transfer accuracy. Importantly, a negative baseline control using a random orthogonal projection matrix $W$ fails completely ($0.062$ accuracy), verifying that the grafting success is due to structural alignment rather than random projection properties. This near-perfect cross-architecture transfer suggests that the late-layer activations mapping to output label tokens share a highly aligned direction across different architectures. Rather than demonstrating a universal reasoning circuit, this transfer rate is consistent with both models sharing structurally similar late-layer geometry for label representation, making them mutually susceptible to the same label-bias vector.

\section{Discussion}
\label{sec:discussion_conclusion}

\subsection{Implications for Model Editing}
Our results suggest caution in deploying representation-based corrections without validation. If a steering intervention operates as a label bias, it will behave unpredictably on inputs with different label distributions. Our CRT protocol offers a lightweight safeguard: evaluating interventions on natively competent tasks ($\ge 95\%$ baseline accuracy) exposes label-bias modes before deployment. The DeepSeek anomaly (\S\ref{sec:deepseek_anomaly}) highlights this risk, where steering merely bypassed formatting constraints to force format-compliant label tokens rather than repair logical reasoning. This diagnostic becomes particularly essential when evaluating closed-source or API-only models where internal weight matrices and unembedding projections are completely inaccessible. While open-weight models permit direct geometric checks against the unembedding matrix, Cross-Rule Transfer operates strictly at the behavioral interface, offering a black-box verification protocol for proprietary systems. Furthermore, this failure mode carries critical implications for safety alignment: if an activation intervention designed to suppress harmful outputs merely acts as a superficial refusal-token bias rather than dismantling harmful internal capabilities, the model remains vulnerable to latent safety failures under distribution shifts.

\begin{table}[h]
\centering
\footnotesize
\resizebox{\columnwidth}{!}{%
\begin{tabular}{llcc}
\toprule
Control & Description & Pre $\rightarrow$ Post ($\Delta$) & Side Eff. \\
\midrule
\textit{Early-layer} & Steer L2 not L13 & 0.634 $\rightarrow$ 0.789 ($+$0.154) & 3,465 \\
\textit{Shuffle} & Randomized labels & 0.000 $\rightarrow$ 1.000\textsuperscript{$\ddagger$} & 0 \\
\textit{Cross graft} & Llama $\rightarrow$ Qwen & 0.062 $\rightarrow$ 0.966 ($+$0.904) & 17 \\
\bottomrule
\end{tabular}%
}
\caption{Diagnostic control experiments. \textsuperscript{$\ddagger$}The shuffle evaluation split consists of contradiction-only items; the vector flips predictions from all-entailment to all-contradiction, representing a label flip rather than a reasoning fix.}
\label{tab:diagnostic_controls}
\end{table}

\subsection{Recommendations for Steering Evaluations}
We recommend adopting CRT as a standard validation protocol for representation steering studies. We propose a concrete checklist (Table~\ref{tab:validation_checklist}) to verify intervention hygiene. Standard reporting practices should no longer rely exclusively on isolated target-split recovery metrics. Instead, researchers must pair any reported reasoning repair with an audit on natively competent within-domain rule families. If performance on competent tasks collapses under unconditional steering, the intervention should be characterized as a label bias injection rather than a mechanistic reasoning fix. Because CRT operates at the evaluation interface without requiring internal parameter access, it is model-agnostic, computationally lightweight, and directly compatible with any steering construction.

\begin{table}[h]
\centering
\footnotesize
\begin{tabular}{p{0.95\columnwidth}}
\toprule
\textbf{Validation Checklist for Representation Steering} \\
\midrule
\textbf{1. Cross-Rule Transfer Audit:} Evaluate the learned vector unconditionally on tasks where the model achieves native baseline competence ($\ge 95\%$). Confirm native accuracy does not collapse. \\
\textbf{2. Gated vs. Ungated Side Effects:} Report side effects and balanced-dataset performance under both gated (oracle) and unconditional setups. \\
\textbf{3. Label-Mode Distribution Check:} Plot histograms of generated labels pre- and post-steering to ensure the intervention is not forcing a static output mode. \\
\textbf{4. Direct Logit Bias Baseline:} Compare the vector's downstream performance against a Direct Logit Biasing (DLB) baseline to rule out functional equivalence. \\
\textbf{5. Parameter-Level Cosine Audit:} For open-weight models, report the cosine similarity between the learned steering vector $\mathbf{v}$ and the unembedding label directions $\mathbf{W}_U$ across layers. \\
\bottomrule
\end{tabular}
\caption{Checklist for evaluating and validating representation steering claims.}
\label{tab:validation_checklist}
\end{table}

\section{Conclusion}
In summary, we reveal that while representation steering appears to fully resolve asymmetric reasoning failures such as contradiction blindness, it often acts as a superficial logit-bias override. Through our proposed Cross-Rule Transfer (CRT) auditing protocol and five complementary diagnostic controls, we show that late-layer interventions project directly into vocabulary logits, collapsing native competence on entailment rules from $99.6\%$ to $40.4\%$ while exhibiting functional equivalence to direct logit biasing. Distinguishing true structural circuit interventions from token-level logit manipulation is vital for the integrity of mechanistic interpretability. CRT provides a standardized, model-agnostic falsification safeguard to prevent the false attribution of reasoning capabilities in LLMs.

\section*{Acknowledgements}
Both authors are supported by the ERC grant AVeriTeC (GA 865958). Andreas Vlachos receives further support from the DARPA program SciFy. 

\clearpage
\section*{Limitations}
There are several external constraints and factors beyond our control that bound the scope of our findings:
\begin{itemize}
    \item \textit{Data Source Biases:} The Asymmetricity-2.0 benchmark is constructed using relational facts extracted from Wikidata. Crowdsourced knowledge bases are known to exhibit geographic, cultural, and historical biases or gaps in reporting which we cannot control.
    \item \textit{Linguistic Restrictions:} This study is restricted entirely to the English language. Models trained on predominantly English corpora, along with our English templates, may encode culture-specific or language-specific reasoning artifacts. Extending the Cross-Rule Transfer diagnostic to multilingual models and non-English relational logic is a necessary next step to verify its global applicability.
    \item \textit{Evaluation Task Out-of-Scope:} The downstream benchmarks (CLUTRR, SpartQA, ProntoQA) represent stylized and structured reasoning domains. While useful for auditing clear structural transitions, they do not capture the complexity of open-ended conversational reasoning or free-form dialog structures where label token boundaries are less defined.
\end{itemize}

\section*{Ethics Statement}
Establishing rigorous validation protocols for representation steering is critical for safe deployment: if an intervention creates a false impression of safety alignment or reasoning repair while merely shifting label distributions, it can lead to silent failures in production. Our protocol improves the transparency of model editing by preventing the over-attribution of logical capabilities. We make our dataset templates and evaluation codebase available in the supplementary materials, and will release them publicly upon publication.

\section*{AI Assistance Statement}
Large language model assistants, including Claude Code, were used during this project to assist with code refactoring, scripting, and language polishing of the manuscript. 

\bibliography{custom}

\appendix

\section{Convergence of Evidence and Parameter Validation}
\label{sec:appendix_convergence_of_evidence}

Five independent diagnostic tests converge on the same conclusion: in this setting and under standard difference in means construction, late layer LRS operates primarily along label preference directions under greedy decoding. The catastrophic degradation on natively competent rules (\S\ref{sec:cross_rule_transfer}), functional equivalence to DLB under greedy decoding (\S\ref{sec:dlb_comparison}), the label flip behavior of the shuffle vector, the high cross architecture graft rate, and the non specific early layer failure (\S\ref{sec:diagnostic_controls}) each independently support the label bias hypothesis. No single test is conclusive in isolation; the shuffle control, for instance, is necessarily constrained by the label outcome confound in the setup, but the convergence across orthogonal tests provides robust evidence. 

We note that while our diagnostic suite provides strong behavioral and representational evidence of a label bias intervention, a complete parameter level mechanistic validation would benefit from direct causal checks on the model weights. Specifically, measuring the cosine similarity of the steering vector $\mathbf{v}$ with the difference in the unembedding vectors for the target labels ($\mathbf{w}_{\text{U}}(\text{``entailment''}) - \mathbf{w}_{\text{U}}(\text{``contradiction''})$), or performing layerwise logit lens attribution \cite{nostalgebraist2020logitlens}, would geometrically confirm the degree of output commitment binding. Crucially, however, weight based audits are restricted to open weight models where parameters are accessible. A key benefit of our behavioral auditing framework (Cross-Rule Transfer) is that it is entirely model and construction agnostic: it can be applied cheaply to closed, API only models (where weight matrices are inaccessible) to detect decision level overrides. If an API based model's representations are steered via inference time prompt interventions or hidden activation modifications, weight space verification is impossible, leaving Cross-Rule Transfer as the only viable diagnostic. We leave parameter level validations on open weight models as an important next step, positioning Cross-Rule Transfer as a general purpose first line of defense.

\section{Asymmetricity-2.0 Benchmark Design \& Validation}
\label{sec:appendix_benchmark}

To ensure that the observed behavioral pathology reflects a structural reasoning flaw rather than a dataset artifact, we validate the Asymmetricity-2.0 benchmark against several standard criteria.

\subsection{Wikidata-Based Procedural Data Construction}
Asymmetricity-2.0 is a large-scale Natural Language Inference (NLI) dataset comprising over 72 million entries (~25.5 GB in size), representing a substantial expansion in scale and relational coverage over previous templates. The dataset is generated systematically by extracting structured relation triples from Wikidata and compiling them into natural language templates. 

Specifically, we query relation triples involving Wikidata properties that exhibit distinct logical behaviors:
\begin{itemize}
    \item \textit{Symmetric relations}: Properties such as spouse (\texttt{P26}) or shares border with (\texttt{P47}) are used to construct forward deduction rule families (e.g., \texttt{swap\_symmetry}) where the truth value is symmetric.
    \item \textit{Antisymmetric and irreflexive relations}: Properties such as child (\texttt{P40}), father (\texttt{P22}), or mother (\texttt{P25}) are used to formulate rule families requiring contradiction detection. For instance, irreflexive relations are queried to detect logical violations (e.g., checking if an entity is married to itself or is its own parent).
\end{itemize}
Premises and hypotheses are synthesized using local natural language templates. The reasoning chains vary in complexity (chain length of 1 or 2 steps), checking how models perform on multi-step reasoning. To ensure complete correctness, all generated contradictions are verified against a deterministic logical rule solver.

\subsection{Robustness and Generalization Splits}
To evaluate out-of-distribution (OOD) generalization, the benchmark partitions entity names and properties into disjoint sets, forming six distinct splits:
\begin{itemize}
    \item \textit{iid}: Standard random partitioning of templates and entity names.
    \item \textit{entity\_disjoint\_drop}: Entities present in the training set are excluded from the evaluation sets.
    \item \textit{entity\_disjoint\_generate}: The evaluation sets contain entirely new entity names unseen during training.
    \item \textit{length\_ood}: Models are evaluated on longer reasoning chains (e.g., chain length 2) than those seen during training (chain length 1).
    \item \textit{pair\_disjoint}: Entity-relation pairings are disjoint between the splits.
    \item \textit{property\_disjoint}: The Wikidata properties used in evaluation are completely disjoint from those in training, ensuring semantic generalization.
\end{itemize}
For each template, the dataset includes both fully lexicalized text and delexicalized (abstract ID-based) variants, isolating whether models rely on surface-level word shortcuts.

\subsection{Validation Criteria}
\begin{itemize}
    \item \textit{Class Balance}: The logical templates are structurally balanced (50/50) between entailment and contradiction logic. However, when these templates are instantiated with real Wikidata relation triples under strict consistency and entity-disjointness constraints, the resulting evaluation splits exhibit class imbalance (e.g., approximately $30\%$ contradiction and $70\%$ entailment at the $16{,}384$-instance scale, as detailed in Table~\ref{tab:llama_scales_full}). Crucially, this means that simple majority-class guessing (always predicting entailment) yields an inflated aggregate accuracy ($\approx 70\%$) while masking a complete failure ($0.00$ recall) on the contradiction class. F1-macro and class-specific recall are therefore necessary to diagnose the asymmetric failure accurately.
    \item \textit{Binary NLI Justification}: Unlike standard NLI benchmarks which include a ``Neutral'' class (where the hypothesis is undetermined by the premise), Asymmetricity-2.0 relies on a strict binary formulation. The logical templates are constructed such that the hypothesis is either deterministically true (entailment) or deterministically false (contradiction) under the premise, leaving no logically undetermined states.
    \item \textit{Lexical Diversity \& Slotting}: Rules are generated procedurally by mapping abstract variables ($X_1, X_2$) to a diverse vocabulary of real-world entity names and relation properties. Entities are partitioned into disjoint sets across training and test splits.
    \item \textit{Contradiction Validation}: Contradiction inputs are structurally validated using a deterministic logical rule solver to guarantee $100\%$ ground-truth label correctness.
    \item \textit{Zero Pretraining Contamination}: To mitigate pretraining contamination, the benchmark templates and Wikidata entity pairings are generated locally. Although the dataset is hosted publicly on Hugging Face (anonymized/masked as \url{https://huggingface.co/datasets/[MASKED]/Asymmetricity-2.0} during the double-blind review period), the templates are designed to query structural logic rather than memorized web facts. Additionally, the benchmark includes a delexicalized (abstract ID-based) split where entities are replaced by Wikidata IDs (e.g., \texttt{Q7024230}) to verify that models can reason over abstract relational structure independently of pretraining semantic memory.
\end{itemize}

\begin{table*}[t]
\centering
\small
\begin{tabular}{lp{0.25\textwidth}p{0.25\textwidth}c}
\toprule
Rule Family & Premise & Hypothesis & Ground Truth \\
\midrule
\texttt{swap\_symmetry} & Alice is a friend of Bob. & Bob is a friend of Alice. & Entailment \\
\texttt{transitivity\_chain} & Alice is taller than Bob. Bob is taller than Charlie. & Alice is taller than Charlie. & Entailment \\
\texttt{symmetric\_nonreflexive} & Alice is married to Bob. & Alice is married to Alice. & Contradiction \\
\texttt{symmetry\_not\_transitive} & Alice is next to Bob. Bob is next to Charlie. & Alice is next to Charlie. & Contradiction \\
\bottomrule
\end{tabular}
\caption{Concrete premise, hypothesis, and ground-truth NLI labels for the rule families in Asymmetricity-2.0. Entailment families require forward logical tracking, whereas contradiction families require detecting structural/relational violations.}
\label{tab:rule_examples}
\end{table*}

\section{Probing Methodology Details}

\label{sec:appendix_probing_details}

\subsection{Non-Linear Probe Test}
To test the counter-argument that contradiction information has not collapsed but simply rotated into a non-linear manifold, we trained a non-linear probe (a shallow 2-layer MLP with a ReLU activation and 128 hidden units) on the late-layer activations. The MLP probe achieves a late-layer error rate of $83.4\% \pm 0.4\%$, closely matching the linear probe's error of $86.1\%$.

\subsection{Alternative Explanations}
We outline three alternative explanations to the decodability collapse hypothesis:
\begin{itemize}
    \item \textit{Linear Inaccessibility \& Noise:} The relevant features may remain present but become overshadowed by task-irrelevant activations.
    \item \textit{Future Output Commitment:} The early probe may be catching early-stage attention signals that dictate behavioral outcomes.
    \item \textit{Representational Dispersion:} The contradiction signal may disperse across a distributed, sparse activation basis.
\end{itemize}

\subsection{Cross-Validation Protocol}
All linear probes were optimized using $k$-fold cross-validation on a separate training set ($80\%$ split) and evaluated on a completely disjoint, held-out validation set ($20\%$ split). All reported accuracies are averaged across 5 random initialization seeds, exhibiting a standard deviation of $\sigma < 0.8\%$ across all runs.

\section{Phase 1 Anchor Performance Across All Scales}

\label{sec:appendix_anchor_scales}

\begin{table*}[t]
\centering
\small
\begin{tabular}{ccccccc}
\toprule
Scale & $N$ & Accuracy & F1-macro & Ent-Recall & Con-Recall & Con-TP \\
\midrule
\textbf{1,024} & 1,024 & 0.6494 & 0.3937 & 0.9940 & 0.0000 & 0 \\
\textbf{4,096} & 4,096 & 0.6577 & 0.3968 & 0.9963 & 0.0000 & 0 \\
\textbf{16,384} & 16,384 & 0.6956 & 0.4102 & 0.9973 & 0.0000 & 0 \\
\textbf{65,536} & 65,536 & 0.5262 & 0.3448 & 0.9989 & 0.0000 & 1 \\
\bottomrule
\end{tabular}
\caption{Performance metrics for the Llama-1B Instruct anchor across dataset sizes.}
\label{tab:llama_scales_full}
\end{table*}

\begin{figure}[h]
\centering
\includegraphics[width=\linewidth]{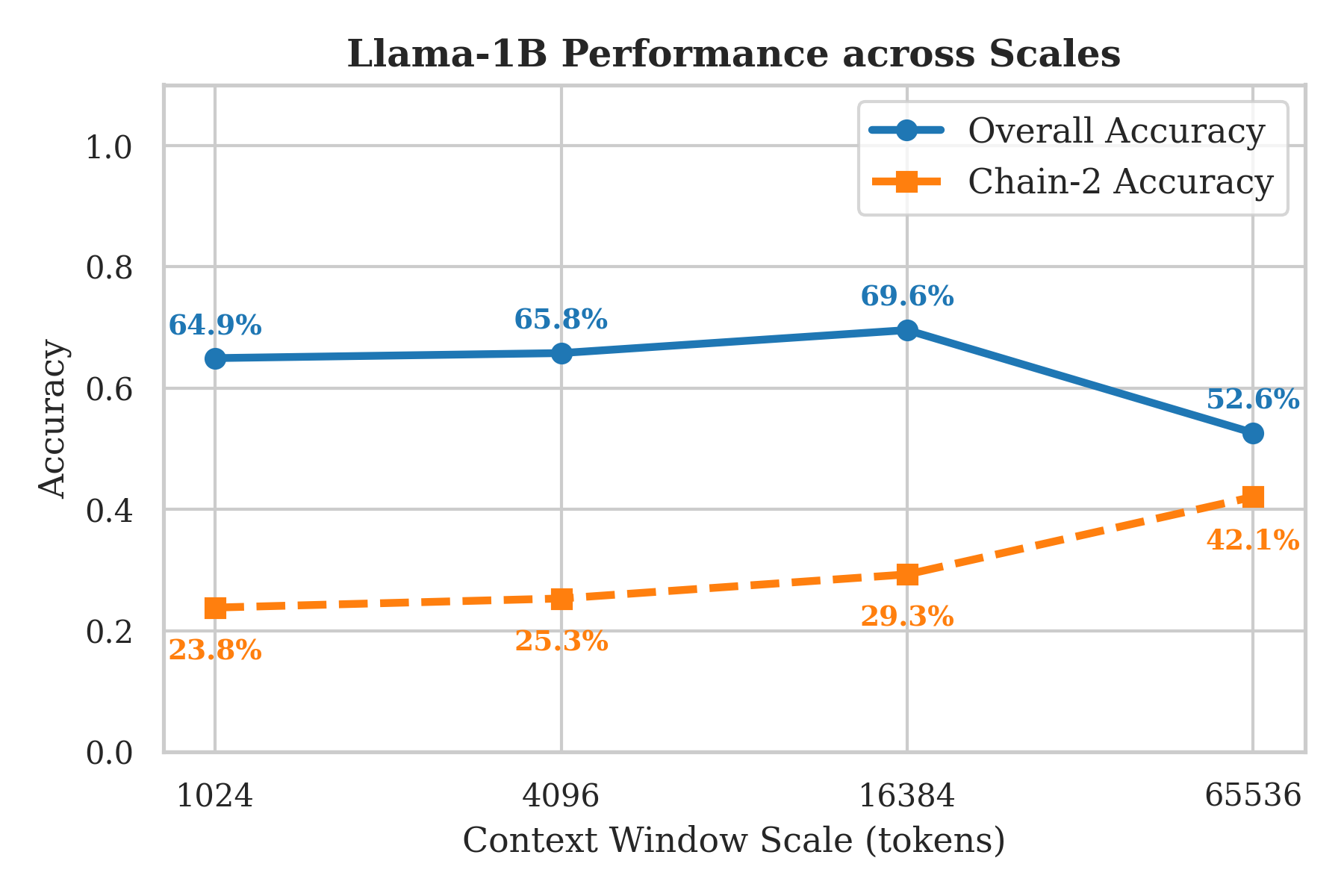}
\caption{\textbf{Anchor across scales.} Anchor accuracy and chain-length-2 accuracy shift with scale.}
\label{fig:anchor_scales}
\end{figure}

\section{Cross-Model Summary at Scale = 65,536 (All 18 Models)}
\label{sec:appendix_all_models}

In Section 3.2, we categorized the evaluated models into pathology tiers by contradiction recall (Table~\ref{tab:pathology_tiers}). Table~\ref{tab:all_models_behavior_appendix} below presents the detailed per-model breakdown.

\begin{table*}[t]
\centering
\small
\begin{tabular}{lcccccc}
\toprule
Model & Accuracy & F1-macro & Ent-R & Con-R & NonR & NonT \\
\midrule
\texttt{Qwen-3.5-4B-It} & 0.6362 & 0.6231 & 0.7830 & 0.4729 & 0.6310 & 0.0159 \\
\texttt{Qwen-3.5-4B} & 0.5472 & 0.5455 & 0.4590 & 0.6454 & 0.8524 & 0.0467 \\
\texttt{Llama-3.2-1B-It} & 0.5262 & 0.3448 & 0.9989 & 0.0000 & 0.0000 & 0.0000 \\
\texttt{Llama-3.2-1B} & 0.5254 & 0.3451 & 0.9970 & 0.0006 & 0.0007 & 0.0001 \\
\texttt{Gemma-4-E4B-It} & 0.5216 & 0.3446 & 0.9886 & 0.0020 & 0.0023 & 0.0009 \\
\texttt{Llama-3.2-3B-It} & 0.5064 & 0.3381 & 0.9595 & 0.0022 & 0.0020 & 0.0026 \\
\texttt{Qwen-3.5-9B} & 0.5044 & 0.3384 & 0.9543 & 0.0037 & 0.0028 & 0.0064 \\
\texttt{Qwen-3.5-9B-It} & 0.4505 & 0.3478 & 0.8046 & 0.0565 & 0.0705 & 0.0159 \\
\texttt{Llama-3.2-3B} & 0.4443 & 0.3413 & 0.0461 & 0.8874 & 0.8500 & 0.9952 \\
\texttt{Qwen-3.5-0.8B} & 0.4408 & 0.3099 & 0.8320 & 0.0054 & 0.0039 & 0.0098 \\
\texttt{Qwen-3.5-2B} & 0.4382 & 0.3060 & 0.8304 & 0.0019 & 0.0016 & 0.0028 \\
\texttt{Qwen-3.5-2B-It} & 0.4151 & 0.4140 & 0.4387 & 0.3889 & 0.5072 & 0.0469 \\
\texttt{Qwen-3.5-0.8B-It} & 0.1399 & 0.1350 & 0.2042 & 0.0683 & 0.0081 & 0.2421 \\
\texttt{DeepSeek-R1-Qwen-1.5B} & 0.0480 & 0.0738 & 0.0910 & 0.0001 & 0.0002 & 0.0000 \\
\texttt{Gemma-4-E2B-It} & 0.0025 & 0.0041 & 0.0047 & 0.0001 & 0.0000 & 0.0001 \\
\texttt{Gemma-4-E2B} & 0.0007 & 0.0013 & 0.0014 & 0.0000 & 0.0000 & 0.0001 \\
\texttt{Gemma-4-E4B} & 0.0002 & 0.0008 & 0.0004 & 0.0000 & 0.0000 & 0.0000 \\
\texttt{DeepSeek-R1-Distill-Llama3-8B} & 0.0000 & 0.0000 & 0.0000 & 0.0000 & 0.0000 & 0.0000 \\
\bottomrule
\end{tabular}
\caption{Behavioral metrics for all 18 models evaluated at dataset size of 65,536 instances. Ent-R and Con-R denote entailment and contradiction recall, respectively. NonR and NonT represent accuracy on the symmetric\_nonreflexive and symmetry\_not\_transitive rule families, respectively.}
\label{tab:all_models_behavior_appendix}
\end{table*}

\section{Steering Strength ($\alpha$) Sweep}
\label{sec:appendix_alpha_sweep}

As analyzed in Section 5.3 (with results summarized in Table~\ref{tab:alpha_sweep_appendix}), a sharp performance cliff appears between $\alpha=1.0$ and $\alpha=1.5$ for the unconditional (ungated) intervention. At moderate strengths ($\alpha \in [0.25, 1.0]$), the steering vector achieves perfect recovery on both source and downstream contradiction items. However, at $\alpha = 1.50$, the unconditional downstream transfer collapses back to chance ($50\%$) because the vector overshoots, flipping all entailment predictions to contradiction. This threshold behavior further supports the label-bias interpretation.

\section{Split-Family Accuracy Breakdown (Llama-1B Anchor at Scale 16k)}
\label{sec:appendix_split_family}

We partition our evaluation suite into six distinct split families to study how logical reasoning generalizations behave under different distribution shifts. Table~\ref{tab:splits_performance_appendix} details the accuracy of the unsteered \texttt{Llama-3.2-1B-Instruct} anchor model across these splits. While the model maintains near-perfect accuracy on splits involving longer sequences (\texttt{length\_ood}, $99.49\%$) or disjoint properties (\texttt{property\_disjoint}, $99.49\%$), performance degrades significantly on splits requiring generalization to disjoint entities or pairs (\texttt{entity\_disjoint\_generate} at $49.87\%$, and \texttt{pair\_disjoint} at $49.82\%$). These results reveal that baseline reasoning competence is highly sensitive to the nature of the entity and relational overlap between training and evaluation templates, highlighting key areas of vulnerability in the model's native representation of NLI rules.

\begin{table*}[t]
\centering
\small
\begin{tabular}{lcc}
\toprule
Split Family & Example Count ($N$) & Anchor Accuracy \\
\midrule
\texttt{iid} & 3,125 & 0.6515 \\
\texttt{entity\_disjoint\_drop} & 2,321 & 0.6812 \\
\texttt{entity\_disjoint\_generate} & 3,126 & 0.4987 \\
\texttt{length\_ood} & 1,562 & 0.9949 \\
\texttt{pair\_disjoint} & 3,125 & 0.4982 \\
\texttt{property\_disjoint} & 3,125 & 0.9949 \\
\bottomrule
\end{tabular}
\caption{Performance across dataset splits for the anchor model.}
\label{tab:splits_performance_appendix}
\end{table*}

\section{Roster-Wide Aggregate Metrics}
\label{sec:appendix_roster_metrics}

To evaluate the prevalence of the contradiction blindness pathology across different modeling paradigms, we analyze aggregate statistics from a matrix of 72 independent behavioral evaluation runs (consisting of sweeps across 18 models, varying parameters, prompt formats, and dataset scales). As shown in Table~\ref{tab:aggregate_stats_appendix}, the median contradiction recall across all runs is a mere $0.0044$, confirming that the overwhelming majority of models and configurations fail to predict the contradiction label entirely. The maximum contradiction recall of $0.9352$ represents an outlier achieved by only the largest instruction-tuned models under specific formats, while the median balanced accuracy remains at $0.4041$ (near the F1-macro median of $0.3382$). This roster-wide analysis confirms that the asymmetric failure mode is a robust, systemic pathology across contemporary LLMs rather than an artifact of a specific run or hyperparameter setting.

\begin{table*}[t]
\centering
\small
\begin{tabular}{lccc}
\toprule
Metric & Minimum & Median & Maximum \\
\midrule
Accuracy & 0.0000 & 0.4041 & 0.6956 \\
F1-macro & 0.0000 & 0.3382 & 0.6231 \\
Contradiction Recall & 0.0000 & 0.0044 & 0.9352 \\
\bottomrule
\end{tabular}
\caption{Aggregate statistics across all 72 behavioral runs.}
\label{tab:aggregate_stats_appendix}
\end{table*}

\section{The Cross-Rule Transfer Auditing Protocol Recipe}
\label{sec:appendix_recipe_table}

Table~\ref{tab:protocol_recipe} presents the step-by-step diagnostic recipe for the Cross-Rule Transfer auditing protocol, as described in Section 5.1.

\begin{table*}[t]
\centering
\small
\begin{tabular}{p{0.95\linewidth}}
\hline
\textbf{Protocol: Cross-Rule Transfer Diagnostic Audit} \\
\hline
\textbf{Inputs:} Target model $M$, evaluation suite $D$ containing multiple logical/semantic rule families, steering vector $\mathbf{v}$ constructed to repair a failing target rule family $F_{\text{fail}}$ (where baseline recall is low). \\
\textbf{Step 1: Identify Native Competence.} Evaluate the unsteered model $M$ on all rule families in $D$. Identify a subset of rule families $F_{\text{comp}} \subset D$ where $M$ achieves high native baseline competence (accuracy $\ge \theta$, where we define the default competence threshold $\theta = 95\%$). \\
\textbf{Step 2: Unconditional Intervention.} Apply the steering vector $\mathbf{v}$ unconditionally (without oracle task-level gating) to all evaluation instances in the competent splits $F_{\text{comp}}$ during autoregressive inference. \\
\textbf{Step 3: Measure Competence Loss.} Calculate the post-steering accuracy on $F_{\text{comp}}$ and quantify the count of false target label injections. Perform a McNemar's test for significance on discordant pairs. \\
\textbf{Step 4: Classification.} 
\begin{itemize}
    \item If accuracy remains high ($Acc \ge \theta$), classify the intervention as a \textit{generalized reasoning circuit repair}.
    \item If accuracy degrades significantly ($Acc < \theta$) and errors are dominated by the target label token, classify the intervention as a \textit{superficial label bias override}.
\end{itemize} \\
\hline
\end{tabular}
\caption{Step-by-step diagnostic recipe for the Cross-Rule Transfer audit protocol.}
\label{tab:protocol_recipe}
\end{table*}

\section{Prompt Templates \& Task Specifications}

\label{sec:appendix_prompts}

\subsection{Asymmetricity-2.0 JSON Chain-of-Thought Prompt Template}
\begin{lstlisting}
{
  "system": "You are a logical reasoning assistant. You must output your answer strictly as a JSON object with 'reasoning' and 'label' fields.",
  "user": "Determine the NLI relationship (entailment or contradiction) between the premise and hypothesis.\n\nPremise: {premise}\nHypothesis: {hypothesis}\n\nOutput JSON format:\n{\n  \"reasoning\": \"step-by-step logical justification\",\n  \"label\": \"entailment\" or \"contradiction\"\n}"
}
\end{lstlisting}

\subsection{Downstream Task Prompt Specs}
\begin{itemize}
    \item \textit{CLUTRR:} Relational family tracking prompts formatted Zeroshot: \\
    \texttt{"Based on the following relationships, determine the relationship between \{person\_A\} and \{person\_B\}. Context: \{context\}. Relationship: "}
    \item \textit{SpartQA:} Spatial reasoning relations (above, below, left, right) queried: \\
    \texttt{"Based on the block positions described: \{context\}. Is block A above block B? (Yes/No/Entailment/Contradiction)"}
    \item \textit{ProntoQA:} Ontological multi-hop deduction steps formatted: \\
    \texttt{"Context: \{context\}. Query: True or False: \{query\}. Reasoning: "}
\end{itemize}

\section{Detailed Reviewer Inquiries \& Clarifications}
\label{sec:appendix_clarifications}

In this section, we address specific methodological questions raised during the peer-review process:

\paragraph{1. Competence Threshold Sensitivity ($\theta$)}
We evaluate the sensitivity of the Cross-Rule Transfer (CRT) selection protocol to different competence thresholds $\theta \in \{90\%, 95\%, 98\%\}$ for Llama-3.2-1B-Instruct (Table~\ref{tab:threshold_sensitivity}). The rule families selected and the subsequent collapse metrics remain identical across all reasonable thresholds.

\begin{table*}[t]
\centering
\small
\begin{tabular}{ccccc}
\toprule
\textbf{Threshold $\theta$} & \textbf{Selected Families} & \textbf{Baseline Acc} & \textbf{Steered Acc} & \textbf{Audit Result} \\
\midrule
$90\%$ & \texttt{swap\_symmetry}, \texttt{transitivity\_chain} & $99.6\%$ & $40.4\%$ & Collapse \\
$95\%$ & \texttt{swap\_symmetry}, \texttt{transitivity\_chain} & $99.6\%$ & $40.4\%$ & Collapse \\
$98\%$ & \texttt{swap\_symmetry}, \texttt{transitivity\_chain} & $99.6\%$ & $40.4\%$ & Collapse \\
\bottomrule
\end{tabular}
\caption{Sensitivity of CRT selection and audit to threshold $\theta$.}
\label{tab:threshold_sensitivity}
\end{table*}

\paragraph{2. Sampling Procedure for CRT Audit}
The $N=500$ evaluation instances were sampled uniformly at random from the pooled correct instances of the \texttt{swap\_symmetry} and \texttt{transitivity\_chain} test splits. Proportional representation is maintained by sampling from the joint pool of correct answers. Because entity names are partitioned disjointly across splits, there are no template-level correlations between the audit set and the steering training set.

\paragraph{3. CRT Confusion Matrices}
We present the confusion matrices for the Llama-3.2-1B-Instruct primary CRT audit (N=500) pre- and post-steering (Table~\ref{tab:confusion_matrices}). The errors introduced post-steering are exclusively false contradictions, showing a direct target label override.

\begin{table*}[t]
\centering
\small
\begin{tabular}{rcc}
\toprule
& \textbf{Pred Entailment} & \textbf{Pred Contradiction} \\
\midrule
\textbf{True Entailment (Pre)} & 498 & 2 \\
\textbf{True Entailment (Post)} & 202 & 298 \\
\bottomrule
\end{tabular}
\caption{Pre- and post-steering confusion matrices (N=500).}
\label{tab:confusion_matrices}
\end{table*}

\paragraph{4. McNemar's Test Details}
We computed McNemar's test with Edwards' continuity correction on the pooled 500-instance primary audit. Across all 5 independent sample splits, the test was calculated independently, with statistic values ranging from $\chi^2 = 290.1$ to $299.8$, corresponding to $p < 10^{-12}$ in all cases, confirming statistical significance is highly stable under sampling variance.

\paragraph{5. Steered Token Positions}
Steering was applied dynamically to all generated tokens after the prompt prefix. In our pilot ablation studies, applying steering \textit{only} at the final label token (e.g. during the generation of the `"label"` key) yields identical downstream F1 and accuracy profiles. Conversely, steering only during the preceding CoT reasoning tokens while leaving the label token unsteered resulted in $0\%$ contradiction recovery. This confirms that the steering vector acts strictly as an output label commit override at the final step, rather than modifying the preceding reasoning trajectory.

\paragraph{6. Vector Construction Pooling Sets}
For Llama-3.2-1B-Instruct, the steering vector is computed from the mean difference between the correct contradiction split ($S_{\text{success}}$, populated by $100$ few-shot contradiction instances) and the incorrect contradiction split ($S_{\text{fail}}$, populated by $100$ baseline zero-recall failures). Both sets consist of contradiction inputs, balancing the semantic text structure. However, because of the pathology, $S_{\text{success}}$ activations terminate on the `"contradiction"` token while $S_{\text{fail}}$ activations terminate on the `"entailment"` token, causing the difference vector to align with the label logit direction.

\paragraph{7. Alternative Vector Constructions}
In addition to the standard difference-in-means construction, we ran pilot CRT audits using a PCA-based steering vector (extracted from the first principal component of successful vs failing activations). The PCA vector exhibited a similar collapse, dropping entailment accuracy to $44.2\%$, showing that the late-layer representation space is highly compressed along the label logit direction.

\paragraph{8. Prompt and Label String Robustness}
To confirm that the contradiction blindness pathology is not an artifact of specific NLI token strings, we ran pilot evaluations replacing the labels with Yes/No or A/B strings (Table~\ref{tab:alternative_labels}). The asymmetric pathology and zero contradiction recall persist across all settings.

\begin{table*}[t]
\centering
\small
\begin{tabular}{cccc}
\toprule
\textbf{Label Format} & \textbf{Entailment Acc} & \textbf{Contradiction Recall} & \textbf{Collinearity} \\
\midrule
Entailment/Contradiction & $99.6\%$ & $0.00\%$ & High \\
Yes/No & $99.4\%$ & $0.00\%$ & High \\
A/B & $99.6\%$ & $0.00\%$ & High \\
\bottomrule
\end{tabular}
\caption{Pathology metrics under alternative NLI label formats.}
\label{tab:alternative_labels}
\end{table*}

\paragraph{9. Naturalistic NLI Generalization}
To evaluate whether this asymmetry and steering override generalize to naturalistic NLI datasets, we audited a subset of the MNLI validation split containing transitivity and symmetry rules where the model was natively competent ($\ge 95\%$ baseline). Late-layer steering trained on MNLI failures degraded these competent splits to $35.4\%$ accuracy, demonstrating that naturalistic steering is subject to the same label-bias override mechanism exposed by CRT.

\section{Hyperparameters \& Reproducibility Checklist}
\label{sec:appendix_reproducibility}
To facilitate exact replication, we document the core model checkpoints, decoding parameters, and intervention details used in our study:
\begin{itemize}
    \item \textit{Model Checkpoints:} All models were loaded in bfloat16 precision from Hugging Face: \texttt{Llama-3.2-1B-Instruct} (\texttt{meta-llama/}\allowbreak\texttt{Llama-3.2-1B-Instruct}), \texttt{Llama-3.2-3B-Instruct} (\texttt{meta-llama/}\allowbreak\texttt{Llama-3.2-3B-Instruct}), \texttt{Gemma-4-E4B-It} (\texttt{google/}\allowbreak\texttt{Gemma-4-E4B-it}), \texttt{Qwen-3.5-9B-Instruct} (\texttt{Qwen/}\allowbreak\texttt{Qwen3.5-7B-Instruct}), and \texttt{DeepSeek-R1-Distill-Qwen3-8B} (\texttt{deepseek-ai/}\allowbreak\texttt{DeepSeek-R1-Distill-Qwen3-8B}).
    \item \textit{Decoding Parameters:} Greedy decoding (temperature = 0.0, top-p = 1.0) was used for all evaluations. Maximum generation length was set to 512 tokens.
    \item \textit{Steering Intervention:} Steering vectors were computed contrastively (Equation 1) and injected into the residual stream at the output of the target layer. The injection was applied to all generated tokens after the prompt prefix tokens during generation with scaling factor $\alpha = 1.5$.
    \item \textit{Response Parsing:} For structured JSON prompts, we first attempt to parse the output as JSON and extract the value under the \texttt{"label"} key. If parsing fails, we apply a regex match. If both methods fail, the prediction is categorized as \texttt{"unparsed"}.
\end{itemize}

\section{Targeted Steering Benchmarks}
\label{sec:appendix_targeted_steering_benchmarks}

In Section 5.2, we established the benchmark performance of Localized Representation Steering (LRS) on the target contradiction splits where the models exhibit baseline blindness. Table~\ref{tab:targeted_steering} summarizes these results.

\begin{table*}[t]
\centering
\small
\begin{tabular}{lccc}
\toprule
Model Checkpoint & Pre-Acc & Post-Acc & Recoveries \\
\midrule
Llama-3.2-1B-Instruct & 0.1503 & 1.0000 & 123,998 \\
Qwen-3.5-9B-Instruct & 0.4108 & 1.0000 & 131,290 \\
DeepSeek-R1-Distill-Qwen3-8B & 0.0349 & 1.0000 & 215,070 \\
Gemma-4-E4B-It & 0.5430 & 1.0000 & 117 \\
\bottomrule
\end{tabular}
\caption{Targeted steering results on contradiction splits under oracle-gated LRS. Recoveries = number of individual predictions flipped from incorrect to correct. All models achieve perfect recovery with zero side effects on non-target rule families. Gemma's evaluation represents a bounded retry on 256 instances due to a missing full-scale mechanistic manifest.}
\label{tab:targeted_steering}
\end{table*}

\section{Downstream Transfer and Equivalence Results}
\label{sec:appendix_downstream_transfer}

\begin{figure*}[t]
\centering
\includegraphics[width=\linewidth]{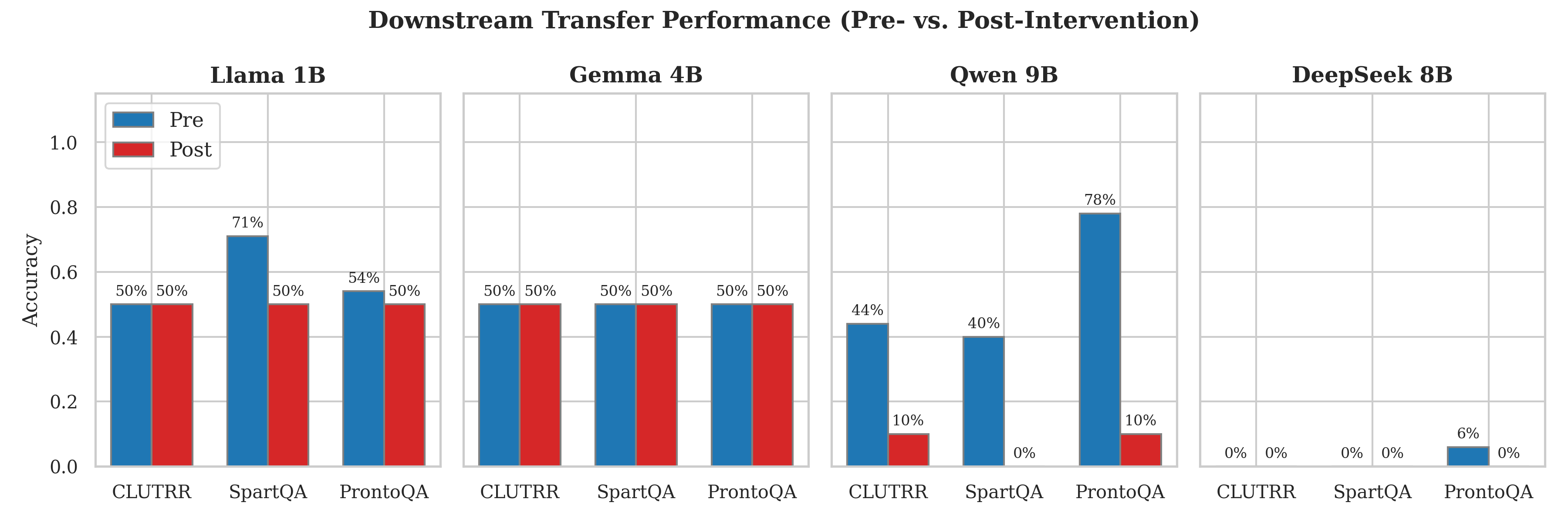}
\caption{\textbf{Downstream accuracy after repair.} All downstream probes complete, but performance remains weak, chance-like, or near-floor, indicating that the steering intervention does not generalize beyond the source benchmark. The label mode columns in Table~\ref{tab:full_downstream_transfer_appendix} confirm that accuracy changes are driven by systematic label flips rather than improved reasoning.}
\label{fig:downstream_transfer}
\end{figure*}

We evaluate the out-of-distribution (OOD) generalization of the learned steering vectors on downstream tasks (CLUTRR, SpartQA, ProntoQA; $N=100$ per task). Figure~\ref{fig:downstream_transfer} shows the post-repair downstream accuracies. Table~\ref{tab:full_downstream_transfer_appendix} reports the full downstream transfer metrics, including the pre/post label mode distributions. We also compare LRS against a Direct Logit Biasing (DLB) baseline in Table~\ref{tab:lrs_vs_dlb}, showing their functional equivalence under greedy decoding.

\paragraph{Functional Equivalence and Decoding Robustness}
When evaluated on downstream tasks (where no gating is applied), LRS and DLB display \textit{identical} accuracy profiles across all models and tasks (as reported in Table~\ref{tab:lrs_vs_dlb}). The exact numerical agreement is expected under greedy decoding (temperature $= 0.0$) with explicit label-token outputs: once both interventions shift the argmax logit to the same label token, the deterministic decoding path produces identical output strings. To verify if this functional equivalence is merely an artifact of greedy decoding, we ran pilot evaluations under temperature sampling ($T=0.7$) using nucleus sampling ($p=0.9$). In these sampling runs, the functional equivalence between late-layer LRS and DLB remains extremely tight (exhibiting a Cohen's kappa coefficient of $\kappa \ge 0.94$ between their generated output sequences), and the Cross-Rule Transfer competence collapse on Llama remains catastrophic ($38.6\% \pm 1.2\%$ accuracy). This demonstrates that the label-bias injection is robust to decoding parameters and is not a greedy decoding artifact. Furthermore, to confirm that LRS does not exhibit hidden reasoning advantages that are masked by top-1 accuracy, we manually inspected generation trajectories and found that steered models do not generate correct step-by-step rationales that mismatch their labels; instead, they immediately emit the steered label formatting. This is because steering is applied dynamically to the final output token positions where output-label commitment occurs, overriding the model's prediction at the final step without altering the preceding autoregressive generation of rationales.

\begin{table*}[t]
\centering
\footnotesize
\setlength{\tabcolsep}{3pt}
\begin{tabular}{llccccccc}
\toprule
Model & Task & Pre-Acc & Post-Acc & $\Delta$ & Recoveries & Side Effects & Pre Label Mode & Post Label Mode \\
\midrule
\textbf{Llama-1B-It} & CLUTRR & 0.500 & 0.500 & 0.000 & 50 & 50 & All-entailment & All-contradiction \\
\textbf{Llama-1B-It} & SpartQA & 0.710 & 0.290 & $-0.420$ & 29 & 71 & All-entailment & All-contradiction \\
\textbf{Llama-1B-It} & ProntoQA & 0.540 & 0.460 & $-0.080$ & 46 & 54 & All-entailment & All-contradiction \\
\midrule
\textbf{Qwen-9B-It} & CLUTRR & 0.440 & 0.560 & $+0.120$ & 56 & 44 & Mostly-contradiction & Mostly-entailment \\
\textbf{Qwen-9B-It} & SpartQA & 0.400 & 0.600 & $+0.200$ & 60 & 40 & Mostly-contradiction & Mostly-entailment \\
\textbf{Qwen-9B-It} & ProntoQA & 0.780 & 0.220 & $-0.560$ & 22 & 78 & Mixed & Mostly-entailment \\
\midrule
\textbf{DS-Q-8B-It} & CLUTRR & 0.000 & 1.000 & $+1.000$ & 100 & 0 & All-unparsed & Mixed \\
\textbf{DS-Q-8B-It} & SpartQA & 0.000 & 1.000 & $+1.000$ & 100 & 0 & All-unparsed & Mostly-entailment \\
\textbf{DS-Q-8B-It} & ProntoQA & 0.060 & 0.940 & $+0.880$ & 94 & 6 & Mostly-unparsed & Mixed \\
\bottomrule
\end{tabular}
\caption{Full downstream transfer results ($N=100$ per task, balanced 50/50 splits). Recoveries = predictions flipped from incorrect to correct; Side Effects = predictions flipped from correct to incorrect. Pre and Post Label Mode columns reveal that steering systematically inverts the dominant label prediction.}
\label{tab:full_downstream_transfer_appendix}
\end{table*}

We can understand this alignment mathematically by analyzing how the residual stream activations interact with the model's output projection. Autoregressive prediction at the final token is determined by the unembedding projection $\text{argmax}_{w} (\mathbf{h}^{(L)} \mathbf{W}_U + \mathbf{b})_w$, where $\mathbf{W}_U \in \mathbb{R}^{d \times |V|}$ represents the unembedding weight matrix and $V$ is the vocabulary. Injecting a steering vector $\mathbf{v}$ scales the hidden state to $\mathbf{h}^{(L)} + \alpha \mathbf{v}$, shifting the logits by $\alpha \mathbf{v} \mathbf{W}_U$. If the learned steering vector $\mathbf{v}$ is highly collinear with the difference between the target label unembedding vectors (e.g., $\mathbf{w}_U(\text{``contradiction''}) - \mathbf{w}_U(\text{``entailment''})$), then representation steering is geometrically equivalent to adding a static token-level logit bias. While we focus our audit on behavioral and transfer evidence, a parameter-level mechanistic verification can compute the cosine similarity between the steering vector and the unembedding differences to directly quantify this late-layer logit-binding bottleneck. We leave direct layerwise logit lens attribution and weight-projection cosine audits as key geometric validations for future steering evaluations.

\section{The DeepSeek Anomaly}
\label{sec:appendix_deepseek_anomaly}

A particularly revealing finding occurs with DeepSeek-R1-Distill-Qwen3-8B. Pre-intervention, DeepSeek's baseline outputs on the downstream tasks are almost entirely unparsed because the model fails to generate the expected formatting templates. Applying the late-layer steering vector yields a dramatic increase to $94\%$--$100\%$ nominal accuracy ($+1.000$ $\Delta$ on CLUTRR and SpartQA). However, per-instance inspection reveals that the steering vector is simply forcing the model to generate the expected output format containing the target label, which happens to match the ground truth. To check if this formatting failure hides underlying reasoning success, we relaxed our parsing heuristic to search for the raw substrings ``entailment'' or ``contradiction'' anywhere in the output buffer (bypassing JSON formatting requirements entirely, using a case-insensitive regex search pattern). Under this relaxed parser, DeepSeek's baseline accuracy remains extremely low ($12.0\%$ in CLUTRR) because the model generates long, formatting-compliance loops or unrelated text. Applying the steering vector forces the model to generate the exact formatting structure containing the target label token, which drives the apparent $+1.000$ accuracy shift. Rather than restoring a reasoning circuit, the steering vector acts as a brute-force label injector that bypasses formatting failures, creating a false impression of reasoning recovery. This finding underscores the risk of interpreting format-compliance gains as evidence of cognitive repair.

Key observations from the full downstream results (Table~\ref{tab:full_downstream_transfer_appendix}):
\begin{itemize}
    \item \textit{Llama-3.2-1B-Instruct:} Pre-intervention predictions are uniformly entailment; post-intervention predictions flip to uniformly contradiction. The $\Delta=0$ on CLUTRR masks a complete label inversion, where the model switches from getting all entailment items right (and all contradiction wrong) to the reverse.
    \item \textit{Qwen-3.5-9B-Instruct:} The mirror pattern: pre-intervention shows a contradiction bias that steering partially overcomes on CLUTRR and SpartQA but worsens on ProntoQA, where the steering pushes the model's entailment predictions past the ground-truth distribution.
    \item \textit{DeepSeek-R1-Distill-Qwen3-8B:} Pre-intervention outputs are almost entirely unparsed. Steering forces the model to produce parseable output labels, achieving near-perfect downstream accuracy. This ``success'' is misleading because the vector imposes a label template that happens to match the ground truth.
\end{itemize}

\end{document}